\def\year{2016}
\documentclass{ecai}
\usepackage{times}
\usepackage{latexsym} 
\usepackage[table, dvipsnames]{xcolor}
\usepackage{booktabs}
\usepackage{graphicx}
\usepackage{multirow}
\usepackage{amsfonts}
\usepackage{tabularx}
\usepackage{tikz}
\usepackage{marginnote}
\usepackage{authblk}

\usetikzlibrary{shapes,snakes}

\begin{document}

\title{Aspect-Based Relational Sentiment Analysis Using a Stacked Neural Network Architecture}
\author {Soufian Jebbara}
\author {Philipp Cimiano}
\affil{Semantic Computing Group,  Cognitive Interaction Technology -- Center of Excellence (CITEC), Bielefeld University, Germany, \{sjebbara,cimiano\}@cit-ec.uni-bielefeld.de} 

\maketitle
\begin{abstract}
Sentiment analysis can be regarded as a relation extraction problem in which the sentiment of some opinion holder towards a certain aspect of a product, theme or event needs to be extracted.
We present a novel neural architecture for sentiment analysis as a relation extraction problem that addresses this problem by dividing it into three subtasks: i) identification of aspect and opinion terms, ii) labeling of opinion terms with a sentiment, and iii) extraction of relations between opinion terms and aspect terms.
For each subtask, we propose a neural network based component and combine all of them into a complete system for relational sentiment analysis.

The component for aspect and opinion term extraction is a hybrid architecture consisting of a recurrent neural network stacked on top of a convolutional neural network.
This approach outperforms a standard convolutional deep neural architecture as well as a recurrent network architecture and performs competitively compared to other methods on two datasets of annotated customer reviews.
To extract sentiments for individual opinion terms, we propose a recurrent architecture in combination with word distance features and achieve promising results, outperforming a majority baseline by 18\% accuracy and providing the first results for the USAGE dataset.
Our relation extraction component outperforms the current state-of-the-art in aspect-opinion relation extraction by 15\% F-Measure.

\end{abstract}

\section{Introduction}
\label{sec:intro}
Sentiment analysis can be regarded as a relation extraction problem in which the sentiment of some opinion holder towards a certain aspect of a product, theme or event needs to be extracted.
While most sentiment analysis methods extract an overall polarity score for a complete text, the following example clearly shows that this is not sufficient:
\\\\
\tikzstyle{every picture}+=[remember picture]
\tikzstyle{none} = [shape=rectangle,inner sep=2pt,outer sep=1pt,text depth=0pt]
\tikzstyle{sentiment} = [shape=rectangle,inner sep=2pt,outer sep=1pt,text depth=7pt]
\tikzstyle{aspect} = [draw,shape=rectangle,inner sep=2pt,outer sep=1pt,text depth=0pt]
\tikzstyle{opinion} = [draw,dashed,line width=0.8pt,shape=rectangle,inner sep=2pt,outer sep=1pt,text depth=0pt]
\tikz\node[none]{The}; \tikz\node[aspect](a1){serrated portion}; \tikz\node[none]{of the}; \tikz\node[aspect](a2){blade}; \tikz\node[none]{is}; \tikz\node[opinion](s1){sharp};\tikz\node[sentiment](sent1){\textit{pos}}; \tikz\node[none]{but the}; \tikz\node[aspect](a3){straight edge}; \tikz\node[none]{was}; \tikz\node[opinion](s2){marginal at best};\tikz\node[sentiment](sent2){\textit{neg}};\tikz\node[none]{.};
\begin{tikzpicture}[overlay]
  \path[->,black,semithick](s1) edge [out=160, in=20] (a1);
  \path[->,black,semithick](s2) edge [out=210, in=-30] (a3);
\end{tikzpicture}
\\\\
The example shows an excerpt of a customer review regarding a kitchen knife.
The opinion expression ``sharp'' constitutes a positive opinion towards the aspect ``serrated portion''.
In the same sentence, a negative opinion is expressed by the phrase ``marginal at best'' towards the aspect ``straight edge''.
Sentiment analysis needs to be regarded thus as a relation extraction problem consisting of three parts:
\begin{enumerate}
\item the extraction of \emph{aspect} and \emph{opinion terms} with respect to the discussed product, theme or event,
\item the labeling of these opinion terms with a \emph{sentiment} (e.g. ``positive'', ``neutral'', ``negative''), and
\item the extraction of \emph{relations} between aspect and opinion terms.
\end{enumerate}

In this work, we propose a complete and modular architecture that addresses all three tasks.
The extraction of aspect and opinion terms can essentially be regarded as a tagging task and can potentially be tackled by sequence modeling techniques such as Hidden Markov Models, Conditional Random Fields (CRFs) etc. Besides, deep neural networks have received increasing interest in recent years and have been applied very successfully to a great variety of natural language processing (NLP)-related tasks. In particular, Convolutional Neural Networks (CNN) have been proposed as a general method for solving sequence tagging problems \cite{collobert2011natural}. Also recurrent neural networks (RNN) have been applied to NLP-related tasks \cite{cho2014learning}.

Building on these encouraging results, in this paper, we employ several neural network based components which we combine into a single architecture to address relational sentiment analysis in three steps.
Firstly, we propose a component that combines convolutional neural networks with recurrent neural networks to extract aspect and opinion terms.
Roughly, the method combines a CNN based sequence tagger with an RNN based tagger by stacking the RNN onto the CNN's produced feature sequence.
The motivation behind this approach is that the RNN on top of the CNN provides a way to preserve information over longer distances in the processed text.
While the deep CNN based tagger is able to capture local dependencies around a word of interest, it cannot incorporate knowledge that appears far away from its current focal position.
The RNN on top, however, might still be able to incorporate locally extracted features of the CNN from far preceding positions in a text through its recurrent hidden layer.

Secondly, a recurrent neural network extracts the expressed sentiment of each opinion term by using Part-of-Speech (POS) tags, word and distance embedding features.
Our method considers the opinion terms in a wide context while still being able to label multiple opinion terms in a single sentence.

\begin{figure}
\centering
  \includegraphics[width=0.8\columnwidth]{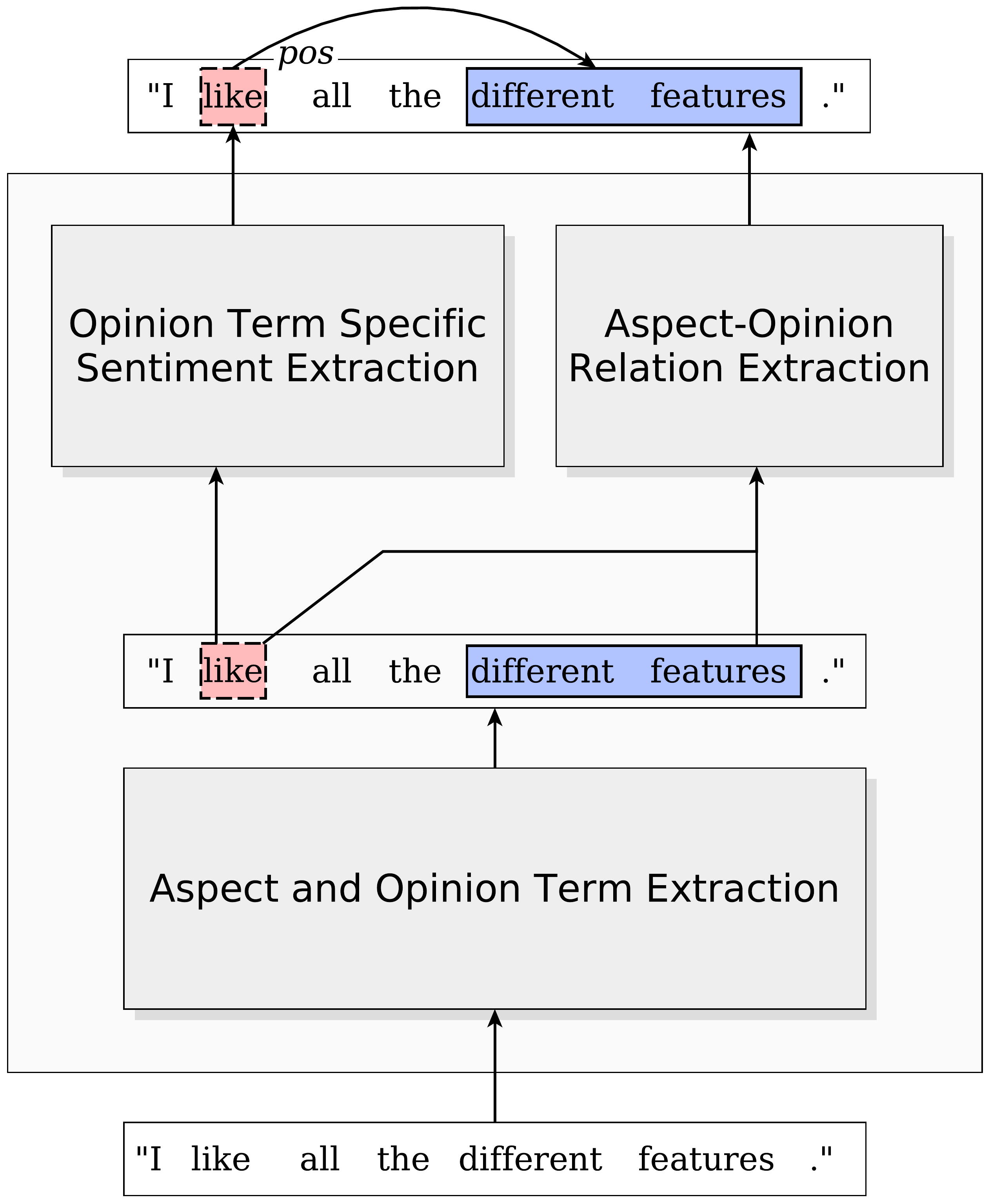}
  \caption{An architecture for sentiment analysis as a relation extraction problem. The architecture comprises 3 components that each address a subtask of the problem: i) identification of aspect and opinion terms, ii) extraction of opinion term specific sentiment and iii) extraction of relations between opinion terms and aspect terms.}
  \label{fig:overview}
\end{figure}

Thirdly, we extract aspect-opinion relations by using a similar RNN model to classify extracted aspect and opinion terms in a pair-wise fashion.
The combination of all these components allows us to realize sentiment analysis on a very fine-grained level, putting individual aspect and opinion mentions in relation.
A schematic visualization of the complete architecture can be seen in Figure \ref{fig:overview}.
Our contributions are the following:
\begin{itemize}
\item We present a complete architecture that addresses sentiment analysis as a relation extraction problem to offer a very fine-grained analysis.
The architecture works in a three step approach by extracting aspect and opinion terms, opinion-term specific sentiment and aspect-opinion relations separately.
\item For all three subtasks, we present a neural network based component that achieves competitive and state-of-the-art results without extensive, task-specific feature engineering.
\item Using the example of aspect and opinion term extraction, we show the impact of training the component with word embeddings initialized from a domain-specific corpus, showing that using domain-specific embeddings increases performance by 6.5\% F-Measure as compared to randomly initialized embeddings.
\item We show the impact of a component performing aspect and opinion term extraction jointly versus predicting each type of phrase separately, demonstrating that joint prediction increases F-measure performance by 1\% for aspect and 5\% for opinion terms.
\item We present a novel approach that is able to extract opinion term specific sentiment.
Our approach achieves performances high above our baseline and provides the first results on the USAGE dataset on the task of opinion term specific sentiment prediction, thus setting a strong baseline for future research.
\item Finally, we show that the relation extraction component is applicable to the sentiment analysis problem and show that our approach outperforms the current state-of-the-art in aspect-opinion relation extraction by 15\% F-measure.
\end{itemize}
The paper is structured as follows:
In the following section, we discuss related work from the domains of aspect based sentiment analysis and relation extraction with respect to the individual subtasks these systems address.
Next, in the sections \ref{sec:aspectopinion}, \ref{sec:sentiment}, and \ref{sec:relation}, we describe the components we use for our three subtasks.
Section \ref{sec:evaluation} describes our evaluation of the individual components that we apply to two datasets.
We describe how parameters of the networks were optimized, the training procedure and the results for the different components on the two datasets.
Lastly, we give a conclusion and discuss issues for future work.

\subsection{Related Work}
Our work is inspired by different related approaches.
Overall, our work is in line with the growing interest of providing more fine-grained, aspect-based sentiment analysis \cite{Le2014,Klinger2013b,Pontiki2015}, going beyond a mere text classification or regression problem that aims at predicting an overall sentiment for a text.  

Vicente et al. \cite{agerri2015elixa} present a system that addresses opinion target extraction as a sequence labeling problem based on a perceptron algorithm with local features.
The system also implements a sentiment polarity classifier to classify individual opinion targets. The approach uses a window of words around a given opinion target and classifies it with an SVM classifier based on a set of features such as word clusters, Part-of-Speech tags and polarity lexicons.

Aspect and opinion term extraction for sentiment analysis has also been addressed using probabilistic graphical models.
Toh and Wang \cite{Toh2014} for instance propose a Conditional Random Field (CRF) as a sequence labeler that includes a variety of features such as POS tags and dependencies, word clusters and WordNet taxonomies.
Additionally, the authors employ a logistic regression classifier to address aspect term polarity classification.
Klinger and Cimiano \cite{Klinger2013a,Klinger2013b} have modeled the task of joint aspect and opinion term extraction using probabilistic graphical models and rely on Markov Chain Monte Carlo methods for inference.
They have demonstrated the impact of a joint architecture on the task with a strong impact on the extraction of aspect terms, but less so for the extraction of opinion terms.
The approach to semantic role labeling proposed by Fonseca et al. \cite{Fonseca2013} is closely related to our approach to aspect term extraction in that the task is phrased as a sequence tagging problem to which a convolutional neural network is applied.

Most relevant in terms of aspect and opinion term extraction are the works of Liu et al. \cite{liu2015finegrained} and Irsoy and Cardie \cite{ozan2014opinion}.
Liu et al. address the extraction of opinion expressions while Irsoy and Cardie focus on the extraction of opinion targets.
Both approaches frame the respective tasks as a sequence labeling task using RNNs.

Our work is related to other approaches using deep neural network architectures for sentiment analysis.
Lakkaraju et al. \cite{lakkaraju2014aspect} for example present a recursive neural network architecture that is capable of extracting multiple aspect categories\footnote{Here, we distinguish between the terminologies of aspect \emph{category} extraction and aspect \emph{term} extraction:
The set of possible aspect categories is predefined and rather small (e.g. \texttt{Price, Battery, Accessories, Display, Portability, Camera}) while aspect terms can take many shapes (e.g. ``sake menu'', ``wine selection'' or ``French Onion soup'').} and their respective sentiments jointly in one model or separately using two softmax classifiers.

Our approach to relation extraction is inspired by the work of Zeng et al. \cite{Zeng2014}, who address a relation extraction task between pairs of entities using a convolutional neural network architecture. Their approach combines lexical features for both entities with sentence level features learned by a CNN model.

All of the above works address a subtask of relational sentiment analysis as we define it in this work, namely i) aspect and opinion term extraction ii) opinion-term specific sentiment extraction and iii) relation extraction.
However, none of the above publications target \emph{all} subtasks in a single architecture or offers the same degree of granularity in the sentiment analysis.
We are thus the first to propose a neural architecture that addresses all three subtasks within one system.

\section{Datasets}
\label{sec:datasets}
For the evaluation of this work, we employ two datasets that provide annotated reviews for the task of relational sentiment analysis.

\subsection{SemEval2015}
\label{sec:semeval}
The SemEval2015 Task 12 dataset \cite{Pontiki2015} is used to evaluate our systems for the subtask of aspect term extraction.
The dataset provides a collection of reviews from different domains (Restaurant, Laptops, Hotels), annotated on different aspect and sentiment levels.
We only make use of the data from the restaurant domain as it contains annotations for explicitly mentioned aspect terms.
The datasets for the laptop and hotel domains only contain annotations for aspect categories without annotated textual mentions.
Note that we can only use this dataset to evaluate our architecture on the task of extracting aspect terms since the dataset is not annotated with respect to opinion terms, and therefore also without aspect-opinion relations.

\subsection{USAGE}
\label{sec:usage}
The USAGE corpus \cite{Klinger2014} is a collection of annotated English and German Amazon reviews of different product categories.
The annotations include (among others) the mentioned aspect terms, the opinion terms marked with a sentiment and relations between aspect and opinion terms.
We refer to Klinger and Cimiano \cite{Klinger2014} for a more detailed description of the dataset.

We restrict our use of this corpus to the annotations for the English reviews and follow the evaluation procedure based on 10-fold cross-validation and strict matching proposed by Klinger and Cimiano  \cite{Klinger2014}.
This dataset provides annotations for all our subtasks hence we will evaluate all our components on this dataset.

\section{Aspect and Opinion Term Extraction}
\label{sec:aspectopinion}
In this work, we compare different choices for neural network based components on the task of aspect and opinion term extraction.
We interpret the extraction task as a sequence labeling task, similar to other sequence labeling tasks \cite{Toh2014,Fonseca2013} and predict sequences of tags for sequences of words.
We use the IOB2 scheme \cite{tksveenstra99eacl} to represent our aspect and opinion annotations as a sequence of tags.
According to this scheme, each word in our text receives one of 3 tags, namely \textbf{I}, \textbf{O} or \textbf{B} that indicate if the word is at the \textbf{B}eginning, \textbf{I}nside or \textbf{O}utside of an annotation:
\begin{center}
\footnotesize
\def\arraystretch{1.5}
\begin{tabular}{cccccccc}
The & \textbf{sake} & \textbf{menu} & should & not & be & overlooked & !\\
O & B & I & O & O & O & O & O
\end{tabular}
\end{center}
In this example the bold ``\textbf{sake menu}'' is an aspect term annotation that we encoded with the IOB2 scheme.
We decided on encoding aspect term annotations separately from opinion term annotations thus resulting in two separate tag sequences per review.
This procedure seems reasonable, since the USAGE dataset allows overlapping aspect and opinion term annotations.
Encoding them into a single tag sequence would require to use a bigger tag set\footnote{The tag set would need to cover single \emph{and} overlapping annotations using the following tag set: \{I-Aspect, I-Opinion, I-Aspect-Opinion, B-Aspect, B-Opinion, B-Aspect-Opinion, O\}} which we would expect to hinder the learning procedure.

For most experiments, we instantiate two separate models -- one for extracting aspect terms and another for extracting opinion terms -- and train both separately to predict their respective tag sequences.
However, as shown in Section \ref{sec:joint} it is also possible to extract aspect and opinion terms jointly without using a larger tag set.
For the actual sequence labeling, we compare convolutional neural networks, recurrent neural networks and combinations thereof since these are capable of dealing with sequential data.

In the following, we discuss the features used by our systems, which include both Part-Of-Speech tags as well as word embeddings.
The word embeddings are learned from a domain-specific corpus of Amazon reviews using a skip-gram model \cite{Mikolov2013}.
Then, we present our different choices of components that we experimentally examine for the aspect and opinion term extraction subtask.
Here, we describe a convolutional network architecture as well as a recurrent network architecture as baseline systems.
We then describe a stacked architecture that feeds features of multiple convolutional layers to a recurrent layer that, in turn, produces a tag sequence.
As a preview for future work, we also propose a component that extracts aspect and opinion terms jointly.

\subsection{Features}
Distributed word embeddings have proven to be a useful feature in many NLP tasks \cite{collobert2011natural,santos2014learning,Le2014} in that they often encode semantically meaningful information about words \cite{Mikolov2013,santos2014learning}.
In this work, we employ word embeddings which we train on huge amounts of reviews since this data is closely related to our main application domain: relational sentiment analysis of customer reviews.
The corpus includes the dataset of McAuley et al. \cite{McAPanLes15,McATarShiHen15} which consists of $\approx 83$ million reviews from 1996 to 2014.
We train the skip-gram model \cite{Mikolov2013} with hierarchical softmax as it is implemented in the topic modeling library \emph{gensim} \cite{gensim2010}.
All reviews are lowercased and the dimensionality of the word vectors is set to $D_{word}=100$.
Rare words that appear less than 10 times in the corpus are replaced with a special token \texttt{<UNK>}.
This token is later used to represent previously unseen words in order to provide a vector for each word at test time.
The resulting vocabulary contains $\approx 1$ million unique words, which we trim to the 200000 most frequent words.

To quantify the impact of using these domain-specific embeddings, we also compute word embeddings on a domain-independent corpus of Wikipedia articles.
As we show in Section \ref{sec:eval:init}, the domain-specific word embeddings indeed outperform the more general Wikipedia word embeddings.

Additionally to the sequence of word embeddings, we evaluate the use of corresponding Part-Of-Speech tags for each word that we obtain from the Stanford POS tagger \cite{Toutanova2003}.
The tag set contains 45 tags plus one additional ``padding'' tag.
We encode these tags in the One-Hot\footnote{A vector of 0s with a single 1 to represent the specific tag.} encoding, which results in a POS tag feature vector with $D_{pos}=46$ dimensions for each word.

\subsection{Convolutional Neural Network Model}
\label{sec:CNN}
While convolutional neural networks were originally intended for image processing, they have been successfully applied to several natural language processing tasks as well \cite{Poria2015,kim2014convolutional,Santos2015}.
When working with sequences of words, convolutions allow to extract local features around each word.

The CNN component for sequence tagging that we propose is composed of several sequentially applied layers that transform an initial sequence of words (i.e. a review) into a sequence of IOB2 tags.
This sequence of tags encodes predicted aspect and opinion annotations for the given review.
More formally, the process from word sequence to tag sequence can be described as follows:

Given a sequence of arbitrary length $N$ of words:
$$[w]_1^N=\{w_1,\dots, w_N\}$$
that correspond to a vocabulary $V$, our model applies a word embedding layer to each sequence element to retrieve a sequence of word embeddings $u_n\in\mathbb{R}^{D_{word}}$:
$$[u]_1^N=\{u_1,\dots, u_N\}.$$
This is done by treating the embedding matrix $W_{word}\in\mathbb{R}^{D_{word} \times |V|}$ as a lookup table and returning the column vector that corresponds to the respective word index\footnote{When using POS tags as additional features, we concatenate the corresponding one-hot vector $p_n$ to the word embedding $u_n$ 
and use the resulting sequence
$[u']_1^N= \{(u_1 \oplus p_1), \dots, (u_N\oplus p_N)\}$
in the next steps.}.

Then, each window of $l_{conv}$ consecutive vectors\footnote{Since we want to apply the convolution operation to the first and the last element in a sequence, too, we pad the input sequence with vectors of 0s.} in $[u]_1^N$ around $u_{n}$ is convolved into a single vector $h^1_n\in\mathbb{R}^{D_{conv}}$, where $D_{conv}$ specifies the number of feature maps for this convolution.
Precisely, the convolution is performed on the concatenated vector $z_n\in\mathbb{R}^{D_{word} \cdot l_{conv}}$ that we define as:
$$z_n=u_{n-(l_{conv}-1)/2} \oplus \dots \oplus  u_{n+(l_{conv}-1)/2},$$
where $\oplus$ represents the concatenation operation.
The convolution at position $n$ in the sequence is then:
$$h^1_n=\sigma(W_{conv}z_n + b_{conv}),$$
where the kernel matrix $W_{conv}\in\mathbb{R}^{D_{conv}\times D_{word}\cdot l_{conv}}$ and the bias vector $b_{conv}$ are shared across all windows for this convolution.
The function $\sigma$ is an element-wise non-linear activation function such as the rectified linear function $f(x)=max(0, x)$.
The resulting sequence is then:
$$[h^1]_1^N=\{h^1_1,\dots, h^1_N\}.$$
This convolution operation can be applied several times (with different $W_{conv}$, $b_{conv}$, $l_{conv}$, and on the output sequence of the previous convolution) to yield a sequence of more abstract representations:
$$[h^m]_1^N=\{h^m_1,\dots, h^m_N\}.$$
In a last step we apply a standard affine neural network layer with a softmax activation function to each individual sequence element that projects the hidden representation $h^m_n$ to a vector of $D_{IOB2}=3$ probability scores that represent the word's affiliation to the corresponding tags I, O or B.

\begin{figure}
  \includegraphics[width=\columnwidth]{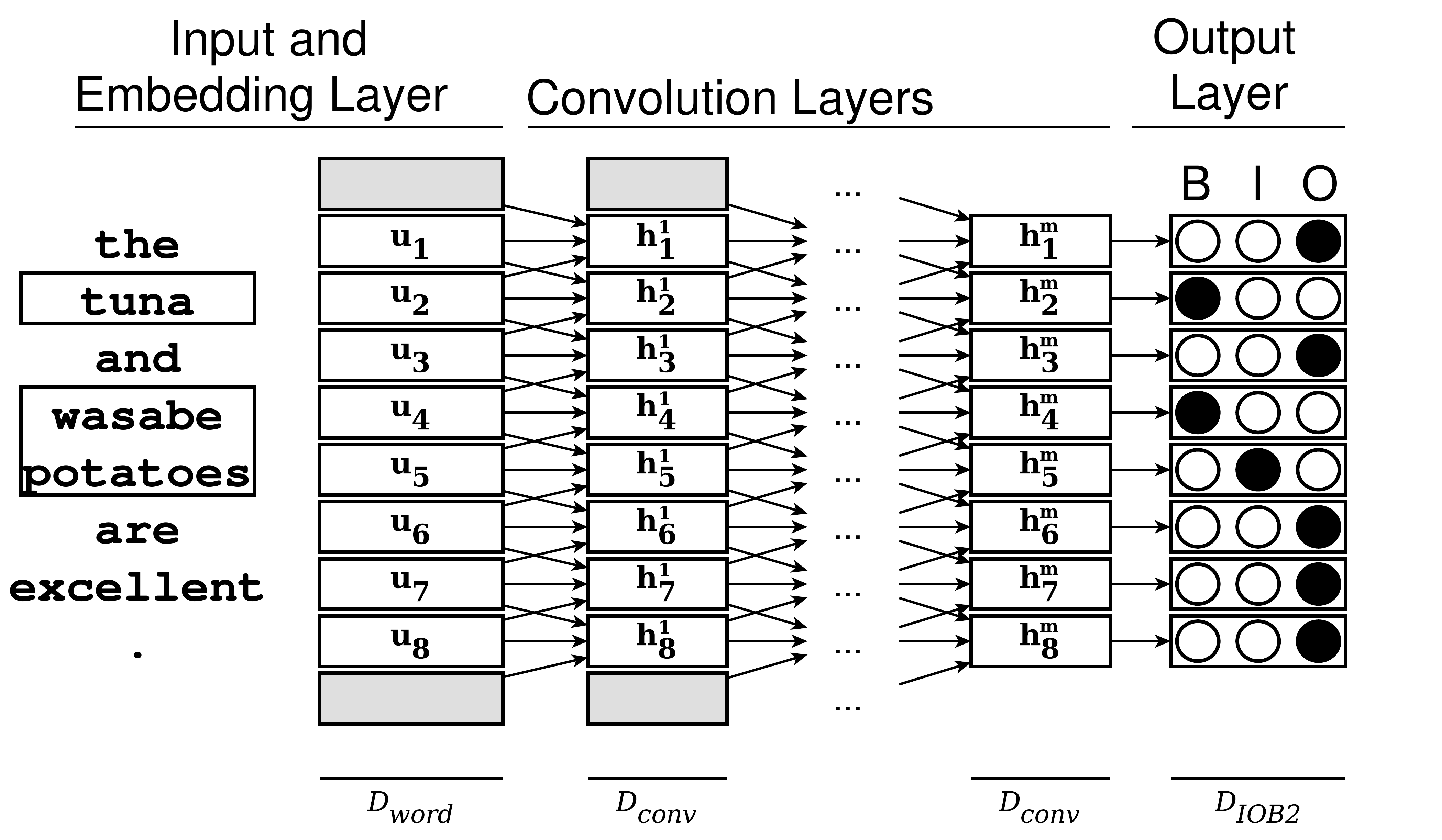}
  \caption{The CNN network for sequence labeling. The words marked with boxes are annotated aspect terms and are tagged with the correct IOB2 tags at the output layer. The gray vectors are padding vectors for the convolution operation.}
  \label{fig:cnn}
\end{figure}

In our following experiments for aspect and opinion term extraction, this CNN architecture consists of a word embedding lookup table for $D_{word}=100$ dimensional vectors, 3 convolution layers and a dense layer that is applied to each single sequence element.
We chose a kernel size $l_{conv} = 3$ and $D_{conv}=50$ feature maps with a rectified linear activation function.
We do not employ a max pooling operation after convolutions in order to retain the initial sequence length.
However, we employ dropout with a drop probability of 0.5 after each convolution as a regularization to prevent overfitting.
The final dense layer uses a softmax activation function to compute probabilities for each of the 3 tags (I, O or B).
The hidden layer sizes for this model are therefore 100-50-50-50 (including the embedding layer).
Figure \ref{fig:cnn} depicts the architecture of this network.
\subsection{Recurrent Neural Network Model}
\label{sec:RNN}
Besides the CNN based sequence tagger, we use an RNN based component to perform aspect and opinion term extraction.
Sequence tagging with an RNN is much more straightforward due to the network's natural handling of sequential input data.
In short, the RNN architecture transforms the input sequence of word indices into a sequence of hidden states using forward and recurrent connections.
Analogously to the CNN approach, the produced hidden states are then mapped to tag probabilities for each of the 3 tags I, O, or B.

For this work, we chose the Gated Recurrent Unit Network (GRU, \cite{cho2014learning}) as the core of our recurrent architecture.
It has been shown that the GRU is a competitive alternative to the well-known Long Short-Term Memory \cite{Hochreiter1997} despite its simpler architecture and less demanding computations \cite{Chung2014}.

Our RNN architecture comprises of an embedding layer for our $D_{word}=100$ dimensional (pretrained) word embeddings, a GRU layer with $D_{gru}=100$ hidden units, and a dense layer with a softmax activation applied to each single output vector of the GRU's output sequence.
The hidden layer sizes for this component are therefore 100-100.

\subsection{Stacked Model}
\label{sec:stack}
The previous two sections describe neural network based models that we consider to tackle aspect and opinion term extraction.
In this section, we propose a combination of the previous two models.

The intuition for combining the CNN and the RNN model is that both models present quite different approaches for the same task.
While the CNN uses locally connected weights to compute localized features around a word of interest, the RNN uses recurrent connections.
The latter connections make it possible to capture important pieces of information from preceding and potentially distant parts of the text.
A combination of both models might benefit from both the local connectivity of the CNN and the recurrence of the RNN.
Similar architecture combinations have been used by Zhou et al. \cite{Zhou2015} and He et al. \cite{He2016}.

We design the new combined model as follows.
First, we apply convolutional layers to the input sequence, similar to the model in \ref{sec:CNN}, yet only up to the final hidden layer.
On top of this sequence of high-level features, we stack a GRU layer that learns temporal dependencies of its input sequence.
Again, a dense layer with a softmax activation is used to map the recurrent hidden states to tag probabilities.
The stacked CNN-RNN model uses the hidden layer sizes 100-50-50-50-100 (or 146-50-50-50-100 when using additional POS tag features).

\subsection{Joint Model}
\label{sec:joint}
Klinger and Cimiano \cite{Klinger2013a,Klinger2013b} present models that are able to extract aspect and opinion terms jointly, leveraging knowledge about aspect terms in order to find opinion terms and vice versa.
Using the stacked model as a basis, we try to replicate this behavior and construct a model that can infer aspects and opinions jointly.

\begin{figure}
  \includegraphics[width=\columnwidth]{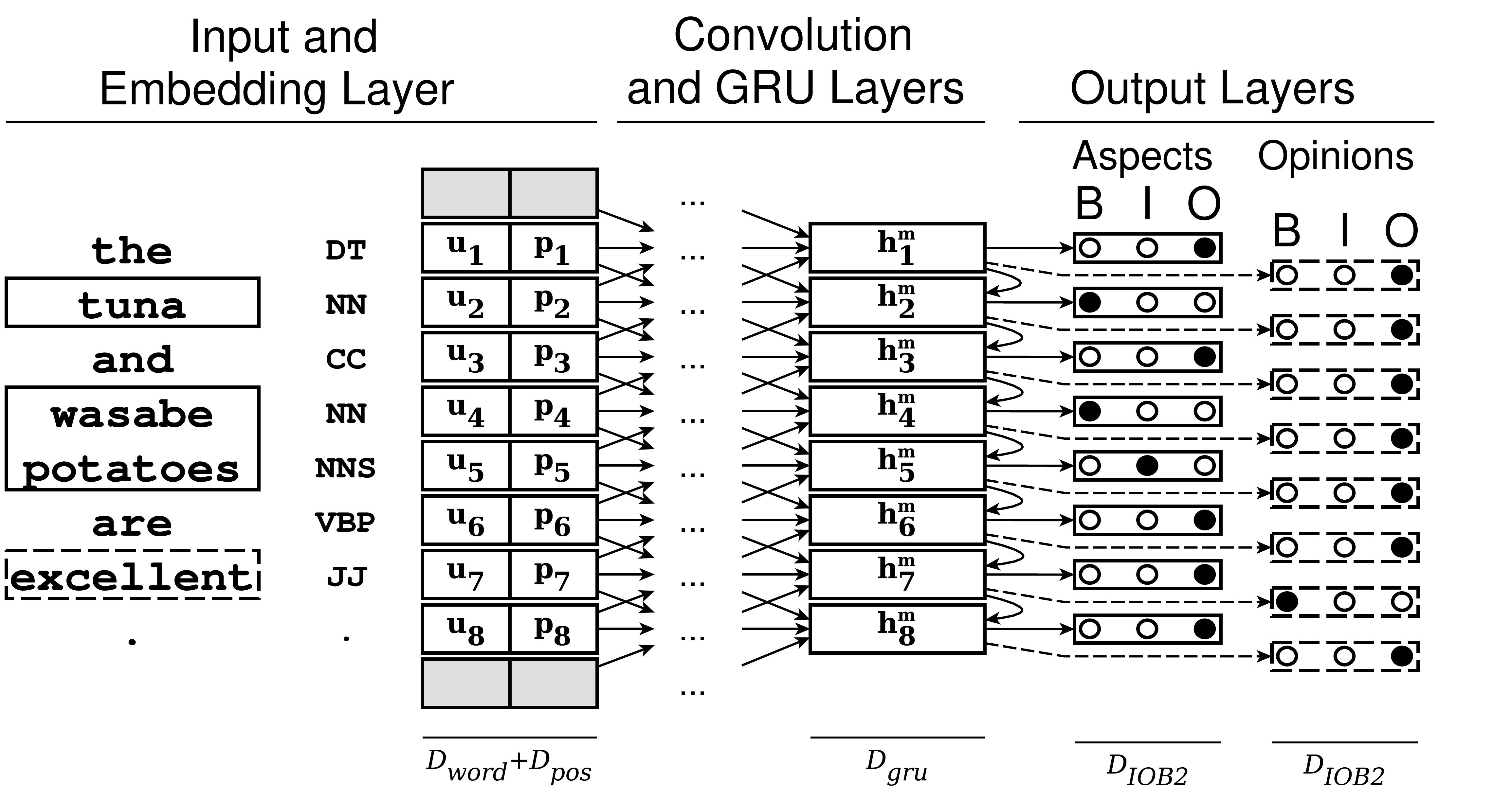}
  \caption{The CNN-RNN network for joint aspect and opinion term extraction. Solid boxes at the input mark aspect terms and are tagged with the correct tags at the aspect output layer (solid arrows and boxes). The dashed box at the input marks an opinion terms. The corresponding output layer is marked with dashed arrows and boxes.}
  \label{fig:joint}
\end{figure}

This model has a very similar structure as the stacked architecture from the previous section.
It differs, however, in a second output layer that we connect to the GRU layer.
With that, the model is able to predict two tag sequences at once: one for extracted aspect terms, the other for opinion terms.
See Figure \ref{fig:joint} for a depiction of the joint architecture.

\section{Opinion Term Specific Sentiment Extraction}
\label{sec:sentiment}
With the previously described models, we are able to detect mentioned aspect and opinion terms, however, without extracting the individual sentiment.
The second step in our pipeline for relational sentiment extraction is to extract sentiments for these opinion terms.
The difficulty here is that it is not enough to extract an overall sentiment for a given review.
Rather, the sentiment needs to be extracted with respect to one of possibly several opinion terms in a text.
In this section, we propose a recurrent neural network architecture together with a combination of word and distance embedding features to address this task.

Given an already detected opinion term in a review, we tag each word in the review with its relative distance to the opinion term in question, as shown in the following example:
\begin{center}
\newcolumntype{C}{>{\centering\arraybackslash}p{0.5cm}}
\footnotesize
\def\arraystretch{1.5}
\begin{tabular}{cccccccc|r}
Coffee & \textbf{stays} & \textbf{fresh} & and & \textit{hot} & in & the & Carafe & (Text)\\
\texttt{NNP} & \texttt{VBZ} & \texttt{JJ} & \texttt{CC} & \texttt{JJ} & \texttt{IN} & \texttt{DT} & \texttt{NN}  & (POS)\\
-1 & 0 & 0 & 1 & 2 & 3 & 4 & 5 & (O)
\end{tabular}
\end{center}
where the bold words ``\textbf{stays fresh}'' form the opinion term for which we want to extract the sentiment.
The italic word ``\textit{hot}'' is another annotated opinion term that is neglected in this extraction step.
Below, the sequence of corresponding POS tags (as obtained with the Stanford POS-tagger) and the relative word distances (O) to the opinion term are shown.

We extract a window of $l_{pol}=20$ words centered around the opinion term instead of considering the whole review text.
Sequences for review texts with less than $l_{pol}$ words are padded at the left with 0s.
Analogously, the sequence of POS tags is extracted.
Using a subsequence of words and POS tags around each opinion term is reasonable since the lengths of the reviews in the USAGE corpus reach up to several hundred words.
Taking the whole review text into account for all opinion terms is computationally very demanding.

We convert each word into its respective vector representation using the pretrained lookup table of word embeddings, thus obtaining a sequence of word vectors.
Similar to this, we also use an embedding layer for the relative distances that provides us with a learned vector representation of dimensionality $D_{dist}=10$ for each distance value, similar to the approach proposed by Zeng et al. \cite{Zeng2014}.
While we did not evaluate those distance embeddings extensively, first results suggest that there is indeed a benefit in mapping the distances to real valued vectors instead of using the raw distance values.
The benefits of distance and position embeddings are supported by the results of other works \cite{Xiang2015,Zeng2014,Sun2015}.

\begin{figure}
  \includegraphics[width=\columnwidth]{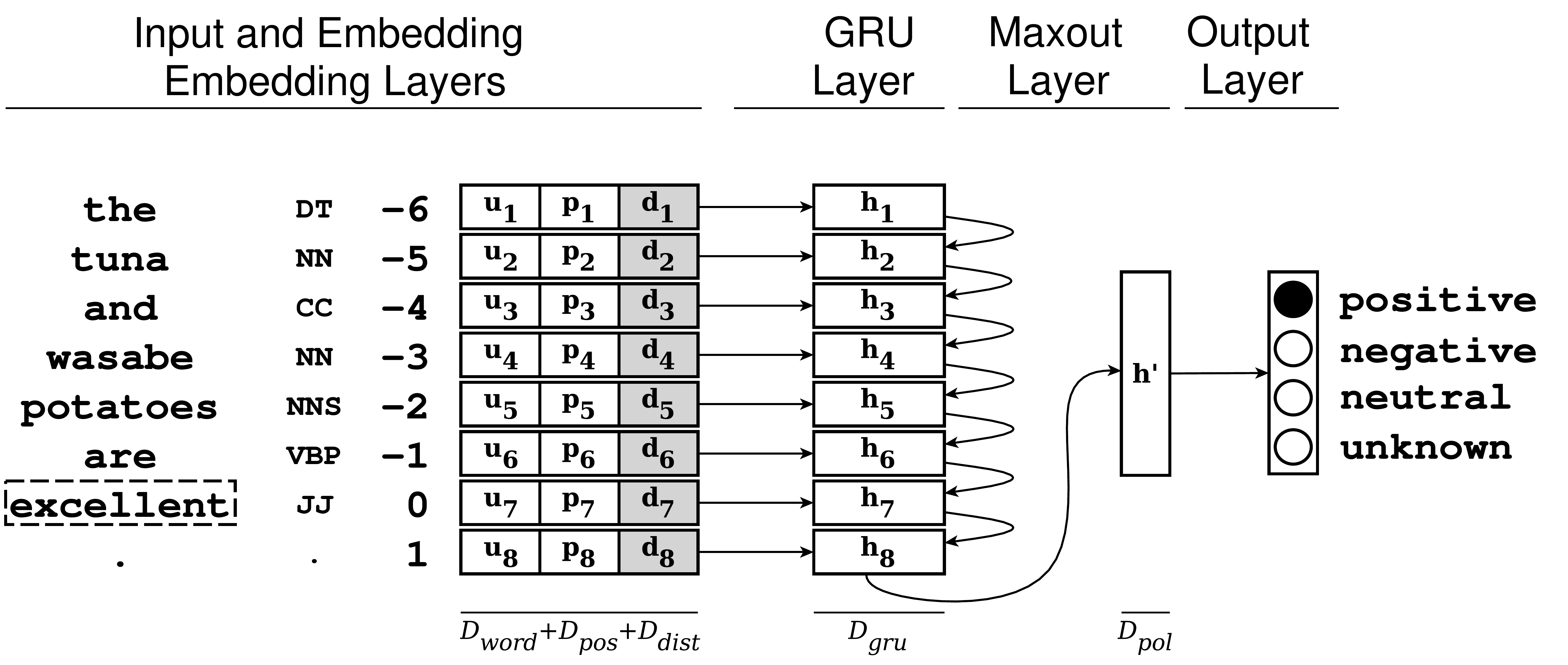}
  \caption{The component for opinion term specific sentiment extraction. The dashed box at the input marks an opinion term. Besides, the corresponding POS tags and the relative word distances are shown. Here, the input vectors are composed of three parts: one vector for the word embeddings, one vector for the POS tag encoding and one vector for the distance embedding. The output layer contains one unit for each possible sentiment label: \texttt{positive},\texttt{neutral}, \texttt{negative} and \texttt{unknown}.}
  \label{fig:sentiment}
\end{figure}

The individual vectors of the three sequences -- word embeddings $u_{n}$, POS tags $p_{n}$ and distance embeddings $d_{n}$ -- are concatenated, resulting in a single sequence of length $l_{pol}$ with $D_{word} + D_{pos} + D_{dist}$ dimensional elements.
We feed the resulting sequence to a recurrent neural network consisting of three layers.
The first hidden layer is a GRU layer with $D_{gru}=100$ hidden units that reads in the sequence of vectors and produces a sequence of hidden states.
The second layer is a densely connected layer of maxout units \cite{Goodfellow2013} that transforms the final hidden state of the GRU layer into a vector $h'$ of $D_{pol}=100$ dimensions.
Lastly, another maxout layer maps the previous hidden layer to 4 output units using a softmax activation function.
Each of the 4 output units corresponds to one of four possible sentiments: \texttt{positive}, \texttt{neutral}, \texttt{negative} and \texttt{unknown}.
The sentiment with the highest probability at the corresponding output unit is the predicted sentiment.
Figure \ref{fig:sentiment} visualizes the network.

\section{Aspect-Opinion Relation Extraction}
\label{sec:relation}
This section introduces the component that is responsible for extracting relations between already extracted aspect and opinion terms.

Besides the annotations for aspect and opinion terms, the USAGE corpus provides annotations of relations between pairs of these aspects and opinions, which we simply refer to as aspect-opinion relations\footnote{The creators of the USAGE corpus actually refer to these relations as \texttt{TARG-SUBJ}. To keep a consistent terminology in this work, we call them aspect-opinion relation.}.
The presence of such a relation implies that the opinion term targets the annotated aspect term.
Due to this binary nature of the relations (\emph{present} or \emph{not-present})  in the USAGE corpus, our relation extraction approach predicts a boolean tag for any given pair of aspect term and opinion phrase.
This pairwise approach allows us to predict the many-to-many relations between aspect and opinion term that are present in the corpus.

Again, we employ a neural network based model that is very similar in structure to the component for opinion term specific sentiment extraction (see previous section).
We adopt a similar strategy as presented in Zeng et al. \cite{Zeng2014} to address relation extraction.
In contrast to Zeng et al. \cite{Zeng2014} however, we use a recurrent neural network instead of a convolutional architecture to perform the actual relation extraction.

Our approach employs four types of features for a given pair of aspect and opinion term:
\begin{itemize}
 \item the sequence of word embeddings of length $l_{rel}=20$ that is centered around the aspect and the opinion,
 \item the sequence of corresponding POS tags for each word,
 \item a sequence of relative distances of each word to the aspect term, and
 \item a sequence of relative distances of each word to the opinion term.
\end{itemize}

As a first step, our model computes the distance (in words) between the two terms and automatically rejects all pairs which are more than 20 words apart from each other.
While this does reject some valid relations, we can still predict $98\%$ of relations correctly for this maximum distance of 20 words.
For those pairs which are below the maximum distance, our model extracts a subsequence of word embeddings of length 20, centered around the aspect term and opinion term.
Sequences for review texts with less than 20 tokens are padded at the left with 0s.
Analogously, the sequence of POS tags is extracted.

To encode information about the distance between the aspect and opinion phrases with respect to their position in the review, we follow a similar approach as Zeng et al. \cite{Zeng2014} and Nogueira dos Santos et al. \cite{Xiang2015}.
We guide the network's attention to the targeted aspect and opinion terms by computing the relative distances of each word in the sequence to the aspect term and to the opinion term, respectively.
\begin{center}
\footnotesize
\def\arraystretch{1.5}
\begin{tabular}{ccccccc|r}
I & \underline{like} & all & the & \textbf{different} & \textbf{features} & . &(Text)\\
\texttt{PRP} & \texttt{VBP} & \texttt{PDT} & \texttt{DT} & \texttt{JJ} & \texttt{NNS} & \texttt{.} & (POS)\\
 -4&-3&-2&-1&0&0&1&(A)\\
-1 & 0& 1&2&3&4&5&(O)
\end{tabular}
\end{center}
Here, the underlined word ``\underline{like}'' marks an opinion term and the bold words ``\textbf{different features}'' mark the aspect term.
Below, the corresponding sequence of POS tags is shown.
The sequences labeled with A and O show the sequence of relative distances of each word to the aspect and opinion term, respectively.

Again, we use an embedding layer to obtain a sequence of $D_{dist}=10$ dimensional embedding vectors for the relative distances.
The individual vectors of the four sequences (word embeddings $u_{n}$, POS tags $p_{n}$, aspect distance embeddings $d_{n}$ and opinion distance embeddings $d'_{n}$) are concatenated, resulting in a single sequence of length $l_{rel}$ and $D_{word} + D_{pos} + 2\cdot D_{dist}$ dimensional elements.

\begin{figure}
  \includegraphics[width=\columnwidth]{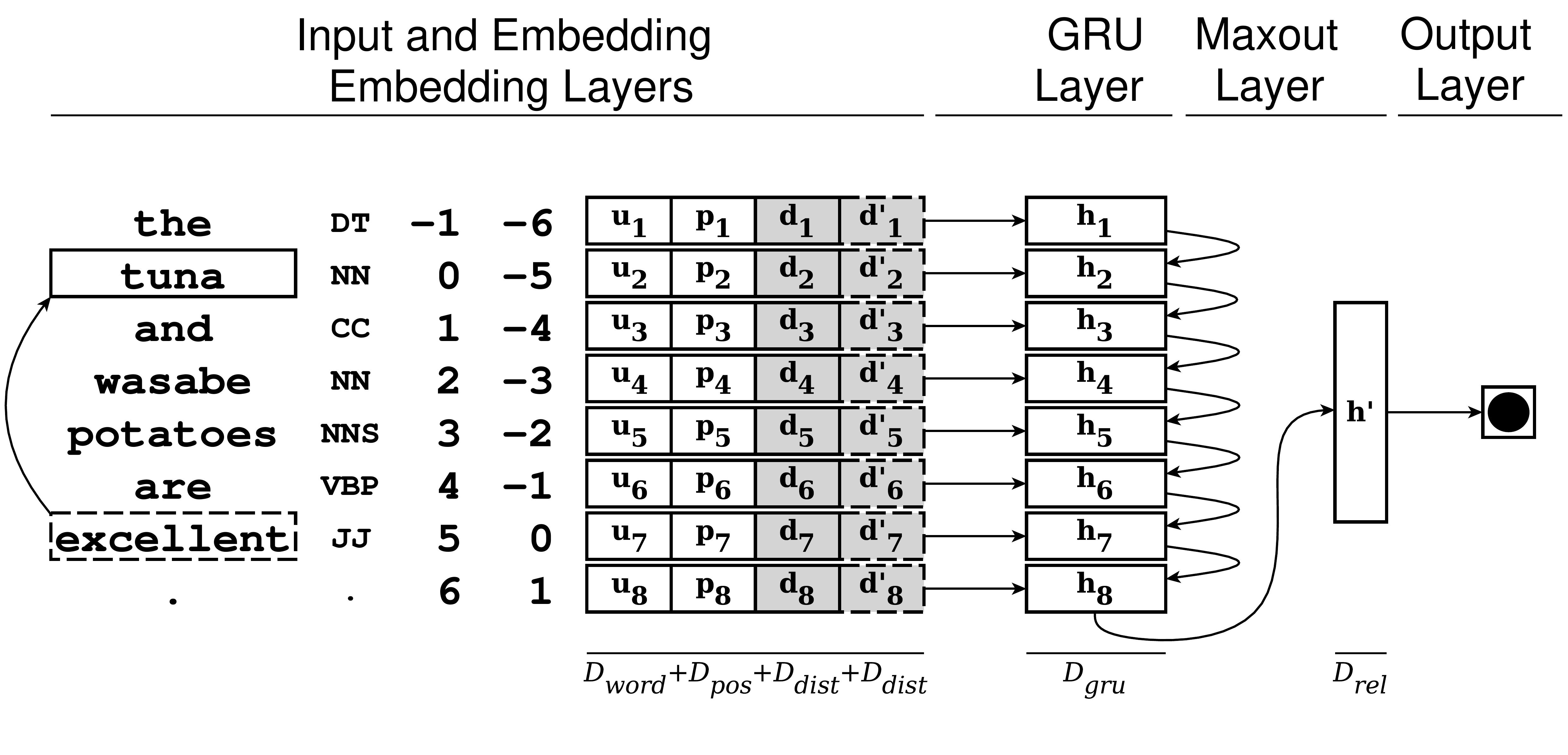}
  \caption{The component for aspect-opinion relation extraction. The solid and dashed boxes at the input mark aspect and opinion terms, respectively. Besides, corresponding POS tags and the relative word distances to the aspect and opinion term are shown. The input vectors are split into four parts: word embeddings, POS tag encoding, aspect distance embeddings and opinion distance embeddings. The output layer contains one single sigmoid unit. An output value $>0.5$ signifies a relation between the aspect and opinion terms.}
  \label{fig:relation}
\end{figure}

We feed the resulting sequence to a GRU layer with $D_{gru}=100$ hidden units that computes a sequence of hidden states.
The final hidden state is passed on to a densely connected layer $h'$ of $D_{rel}=100$ maxout units.
As a last step, the output of the previous hidden layer is passed to a final fully-connected maxout layer with a single output unit and a sigmoid activation function to map its output to a value between 0 and 1.
We interpret the network's output as the probability that the pair of aspect and opinion terms form an aspect-opinion relation.
Figure \ref{fig:relation} visualizes the network.

\section{Experiments and Results}
\label{sec:evaluation}
This section evaluates our proposed architecture for relational sentiment analysis.
Our evaluation targets the individual components of the overall system, measuring their performances in isolation.
For the evaluation, we consider the two datasets described in Section \ref{sec:datasets}.
Both datasets offer a different granularity in their annotations.
The SemEval dataset does not provide annotations for opinion terms, opinion term specific sentiment or aspect-opinion relations.
As such, we only evaluate our component for aspect term extraction on this dataset.
The USAGE dataset, on the other hand, offers annotations for all our subtasks, allowing us to evaluate all our components on this dataset.
Where possible, we give precision, recall and $F_1$ scores for our own and baseline approaches.

All experiments were performed with the deep learning library \emph{Keras} \cite{keras2015} and employ many of its pre-implemented algorithms.
Tag sequences predicted by our approach are post-processed to yield only sequences that are valid according to the IOB2 scheme.

\subsection{Training}
This section briefly outlines our training procedure.
Since we deal with sequences of variable lengths, training our models in mini batches would require to pad shorter sequences with special padding elements.
Due to the large differences in the length of reviews in the USAGE corpus ($18$ to $>2000$ words), we avoid padding sequences with too many padding elements.
Instead, we train our models on one data sample at a time.
The optimization of the models' weights is performed with \emph{RMSProp} \cite{Tieleman2012}, which converges after a few epochs.

Since the performance did not depend strongly on the number of iterations over the training data, we trained all aspect and opinion term extraction models for 15 epochs.
The component for sentiment extraction was trained for 14 epochs and the relation extraction component performed best with 28 epochs of training.

\subsection{Initialization of Word Embeddings}
\label{sec:eval:init}
In our first experiment, we compare the performance impact of initializing the weights of the word embedding lookup table in the stacked CNN-RNN model randomly to the initialization with our pretrained embeddings.
Our intuition is that both sets of pretrained embeddings are helpful for the extraction of aspect and opinion terms but more so the domain-specific embeddings. 
These are expected to capture the most task relevant semantics which might help with the extraction task.

Since many of our network parameters are initialized randomly and our training data is processed in a random order, the performance of our model might be influenced by these external factors.
To mitigate these effects, we performed each experiment three times and averaged the outcomes.

The experiments for aspect term extraction on the SemEval2015 dataset show that our model achieves on average an $F_1$ score of $0.581$ with the randomly initialized embeddings.
The initialization with the Wikipedia embeddings leads to a much higher score, namely $F_1=0.618$, while the domain-specific review embeddings yield an F-Measure of $0.646$.
As expected, we can observe a large benefit in pretraining our word embeddings on huge collections of natural language texts with the largest gain using domain-specific embeddings.

Considering these first results, we only use the domain-specific review embeddings as initialization in further experiments.

\subsection{Evaluation: Aspect and Opinion Term Extraction}
This section evaluates our relational sentiment analysis architecture focusing on the component for aspect and opinion term extraction.
We perform the evaluation in two steps:
First, we evaluate the component on the SemEval dataset measuring its performance for aspect term extraction only.
Keep in mind that we can neither evaluate opinion term extraction nor opinion term specific sentiment analysis or aspect-opinion relation extraction on this particular dataset due to granularity of the provided annotation.
Those subtasks, however, are evaluated in Sections \ref{sec:eval:joint}, \ref{sec:eval:sentiment} and \ref{sec:eval:relation}.

\subsubsection{CNN vs. RNN vs. Stacked Models}
\label{sec:eval:stack}
This section evaluates our aspect term extraction component and considers different neural models.
We train and test the CNN, the RNN and the stacked CNN-RNN model from the sections \ref{sec:CNN}, \ref{sec:RNN} and \ref{sec:stack} on the SemEval2015 dataset using the official training and test split.
Again, we perform each experiment three times to alleviate differences due to the networks' initializations and training sample orders.
Table \ref{tab:models} shows the average precision, recall and $F_{1}$ score for each model.
\begin{table}
  \begin{center}
    \begin{tabular}{lrrr}
      \toprule
      & \multicolumn{3}{c}{Aspects} \\
      \cmidrule(rl){2-4}
      Model & $P$ & $R$ & $F_{1}$ \\
      \midrule
      RNN & 0.592 &  0.646 &  0.618\\
      CNN & 0.558 &  0.702 &  0.621\\
      Stack & 0.599 &  \textbf{0.703} &  0.646\\
      Stack+POS & \textbf{0.633} &  0.689 &  0.659\\
      EliXa &  -- & -- & \textbf{0.701}\\
      \bottomrule
    \end{tabular}
    \caption{Results for aspect term extraction on the SemEval2015 test dataset using different model architectures and additional POS tag features. \emph{EliXa} represents the current state-of-the-art.}
    \label{tab:models}
  \end{center}
\end{table}

The model based on convolutional layers and the model based on recurrent layers both perform similarly regarding the overall $F_{1}$ score.
Combining both types of models in a stack-like architecture results in an increased averaged $F_{1}$ score that is statistically significant with $p=0.05$.
Providing additional features in the form of POS tags does also improve the model's performance.
While we still perform not quite as good as the current state-of-the-art system EliXa \cite{agerri2015elixa} we do perform well with respect to the overall ranking of the SemEval2015 task as can be seen in Pontiki et al. \cite{Pontiki2015}.

In spite of being a few percentage points below the current state-of-the-art system EliXa, our proposed method still constitutes a meaningful contribution.
The benefit of our method is that it is not restricted to the mere extraction of aspect terms but that it is capable of extracting aspect and opinion terms jointly.
The following section shows that our method achieves state-of-the-art performance on the USAGE dataset with this more complex joint extraction.

Based on the results of these experiments, we perform all following aspect and opinion phrase extraction tasks using the CNN-RNN model with additional POS tag features.

\subsubsection{Joint vs. Separate Models}
\label{sec:eval:joint}
This section investigates the benefits of predicting aspect and opinion phrases of the USAGE corpus jointly in one model, in contrast to two separate models.
For this, we consider two joint models with different hidden layer sizes, namely with 100-50-50-50-100 (dubbed \emph{Joint small}) and 100-100-100-100-200 (dubbed \emph{Joint large}).
This should account for differences in performance that solely result from different network capacities.
As proposed by Klinger and Cimiano \cite{Klinger2014}, we evaluate our models by 10-fold cross validation.
Table \ref{tab:joint} shows the results for the joint and the separate models. 
\begin{table}
  \begin{center}
    \begin{tabular}{lrrrrrr}
      \toprule
      & \multicolumn{3}{c}{Aspects} & \multicolumn{3}{c}{Opinions}\\
      \cmidrule(rl){2-4} \cmidrule(rl){5-7}
      Model & $P$ & $R$ & $F_{1}$& $P$ & $R$ & $F_{1}$\\
      \midrule
      Klinger2014 & -- & -- & 0.56 & -- & -- & 0.48 \\
      Aspect only & 0.63 & \textbf{0.70} & 0.66 & -- & -- & -- \\
      Opinion only & -- & -- & -- & 0.44 & 0.48 & 0.45 \\
      Joint small & 0.57 & 0.65 & 0.61 & 0.40 & 0.40 & 0.40 \\
      Joint large & \textbf{0.65} & 0.69 & \textbf{0.67} & \textbf{0.47} & \textbf{0.53} & \textbf{0.50} \\
      \bottomrule
    \end{tabular}
    \caption{Performances for aspect and opinion phrase extraction on the USAGE dataset for joint and separate models. \emph{Joint small} and \emph{Joint large} are the joint models with hidden layer size 100-50-50-50-100 and 100-100-100-100-200, respectively.}
    \label{tab:joint}
  \end{center}
\end{table}

We can see that extracting aspect and opinion terms jointly does indeed enhance the model's performance, but only so for a larger network configuration.
The extraction of opinion terms benefits from the joint setting in particular.
However, this might also be attributed to the increased size of the hidden layers in our neural architecture.
The extraction of opinion terms might simply require more network parameters which it is able to claim in the larger joint architecture.
A more detailed investigation needs to be conducted in future work.

Bear in mind that we do not compare our results on this dataset with the EliXa system referenced in Section \ref{sec:eval:stack}.
The EliXa system is applicable to aspect term extraction in isolation and is not designed for opinion term extraction.

\subsection{Evaluation: Opinion Term Specific Sentiment Extraction}
\label{sec:eval:sentiment}
Next, we show our evaluation of the component proposed in Section \ref{sec:sentiment}.
The model considers individual opinion terms in a wide context and predicts one of four sentiment labels for each presented opinion term.

We perform the sentiment extraction on the opinion terms of the gold standard annotations of the USAGE corpus in order to measure the sentiment extraction in isolation.
To keep the experiments consistent across our different components, we perform the prediction on a 10-fold cross validation of the USAGE documents.
The results in Table \ref{tab:sentiment} show the average accuracy\footnote{We report accuracy since precision, recall and $F_1$ score are all equal in the case where the number of annotations is fixed.} of predicting the sentiment label of an opinion term with the number of correctly and incorrectly labeled opinion terms.
We also show the results of a naive approach that always predicts the (most frequent) sentiment label \texttt{positive} to act as a simple baseline.
\begin{table}
 \begin{center}
    \begin{tabular}{lrrr}
      \toprule
      Model & $Accuracy$ & $\# Correct$ & $\#Incorrect$ \\
      \midrule
      Positive Only & 0.647 & 342.6 & 189.5\\
      Our Approach & \textbf{0.831} & \textbf{441.6} & \textbf{90.5}\\
      \bottomrule
    \end{tabular}
    \caption{$Accuracy$ for opinion term specific sentiment extraction on gold annotations.}
    \label{tab:sentiment}
  \end{center}
\end{table}

We see that our method achieves an accuracy high above our baseline.
Unfortunately, up to date, no results are published for sentiment extraction on the USAGE dataset that we can use as a further baseline.
Hence, with this work, we contribute the first results for opinion term specific sentiment extraction for this dataset.

\subsection{Evaluation: Aspect-Opinion Relation Extraction}
\label{sec:eval:relation}
This part of our evaluation focuses on the last component in the overall architecture for relational sentiment analysis that is responsible for the extraction of aspect-opinion relations.
As before, we perform the relation extraction on the aspect and opinion terms from the gold standard annotations of the USAGE corpus, in order to measure our system's performance for relation extraction in isolation.
This methodology is adopted from Klinger and Cimiano \cite{Klinger2014} and allows us to compare our method to their work.
Table \ref{tab:relation} shows the results of a 10-fold cross-validation of our proposed component.
\begin{table}
  \begin{center}
    \begin{tabular}{lrrr}
      \toprule
      Model & $P$ & $R$ & $F_{1}$ \\
      \midrule
      Klinger2014 & -- & -- & 0.65 \\
      Our Approach & 0.87 & 0.75 & \textbf{0.81} \\
      \bottomrule
    \end{tabular}
    \caption{$F_{1}$ score for aspect-opinion relation extraction on gold annotations.}
    \label{tab:relation}
  \end{center}
\end{table}
The results show that our RNN-based model improves relation extraction by 15\% F-Measure compared to the probabilistic graphical model of proposed by Klinger and Cimiano \cite{Klinger2014}.

\section{Conclusion and Future Work}
In this work, we presented a modular architecture that addresses sentiment analysis as a relation extraction problem.
The proposed architecture divides the problem into three subtasks and addresses each with a dedicated component.
This highly flexible approach offers a fine-grained solution for sentiment analysis.

As part of this overall architecture, we presented possible implementations for the individual components:
First, we presented different neural network models that are capable of aspect and opinion term extraction and which achieved competitive and state-of-the-art results on different datasets.
We could report a benefit for this task in using domain-specific word embeddings compared to domain-independent and randomly initialized embeddings.
We investigated the extraction of aspect and opinion terms separately and jointly and found the joint approach to produce superior results in one setting.
Thus, we confirm previous results from Klinger et al. \cite{Klinger2013a} who used a probabilistic graphical model instead of a neural network model. 

Secondly, we addressed opinion term specific sentiment extraction with a recurrent neural network model and distance embedding features and achieved promising results; the first sentiment extraction results on the considered dataset.

Thirdly, as another contribution, we designed and evaluated a model for the extraction of relations between aspect and opinion terms which outperformed prior results on the same dataset.
Our work shows that it is possible to divide sentiment analysis in a flexible and fine-grained way using a highly modular architecture.

All proposed components stand out by their minimal use of hand-engineered features that are strongly tuned to their specific tasks.
The only external resources that were used are machine-generated POS tags and word embeddings which were created with a data-driven approach.
Nevertheless, all components perform competitive on their individual subtasks.
It is easily conceivable to provide further task-specific features to improve the performances of the individual components even further.
This, however, is not the goal of this work, which is why we leave this part for future work.

Furthermore, we expect our components to benefit from bidirectional recurrent connections \cite{Schuster1997} so that words appearing later in a sentence can be taken into account.
The investigation of attention mechanisms for RNNs \cite{Bahdanau2014} seems also very promising to allow the models to focus more strongly on important parts of the input sequence.

\ack This work was supported by the Cluster of Excellence Cognitive Interaction Technology 'CITEC' (EXC 277) at Bielefeld University, which is funded by the German Research Foundation (DFG).

\bibliographystyle{ecai}
\bibliography{bibliography.bib}

\end{document}